\pdfoutput=1

\documentclass[11pt]{article}

\usepackage{acl}

\usepackage{times}
\usepackage{latexsym}
\usepackage{graphicx}
\usepackage{float}
\usepackage{color, colortbl}
\definecolor{Gray}{gray}{0.9}
\usepackage{graphicx}
\usepackage{makecell}
\usepackage{lipsum}
\usepackage{float}
\usepackage{booktabs}
\usepackage{placeins}
\usepackage{hyperref}
\usepackage{epstopdf}
\usepackage{comment}
\usepackage{algorithm2e}
\usepackage{amsmath}
\usepackage{amssymb}
\usepackage{multirow}
\usepackage{subcaption}

\usepackage[T1]{fontenc}

\usepackage[utf8]{inputenc}

\usepackage{microtype}
\title{Enhancing Adverse Drug Event Detection with Multimodal Dataset: Corpus Creation and Model Development}

\author{
    Pranab Sahoo$^1$,
    Ayush Kumar Singh$^1$,
    Sriparna Saha$^1$,
    Aman Chadha$^{2,3}$\thanks{\,\,\,Work does not relate to position at Amazon.}\,\,, 
    Samrat Mondal$^1$
    \\
    \vspace{0.2cm}
    $^1$Department of Computer Science And Engineering, Indian Institute of Technology Patna\\
    $^2$Stanford University, $^3$Amazon GenAI\\
    \vspace{0.1cm}
    \small
    \begin{tabular}[t]{@{}c@{}}
        \texttt{pranab\_2021cs25@iitp.ac.in, ayush\_2211ai27@iitp.ac.in, sriparna@iitp.ac.in, samrat@iitp.ac.in}\\
        \texttt{hi@aman.ai}
    \end{tabular}
}

\begin{document}
\maketitle
\begin{abstract}
The mining of adverse drug events (ADEs) is pivotal in pharmacovigilance, enhancing patient safety by identifying potential risks associated with medications, facilitating early detection of adverse events, and guiding regulatory decision-making. Traditional ADE detection methods are reliable but slow, not easily adaptable to large-scale operations, and offer limited information. With the exponential increase in data sources like social media content, biomedical literature, and Electronic Medical Records (EMR), extracting relevant ADE-related information from these unstructured texts is imperative. Previous ADE mining studies have focused on text-based methodologies, overlooking visual cues, limiting contextual comprehension, and hindering accurate interpretation. To address this gap, we present a MultiModal Adverse Drug Event (MMADE) detection dataset, merging ADE-related textual information with visual aids. Additionally, we introduce a framework that leverages the capabilities of LLMs and VLMs for ADE detection by generating detailed descriptions of medical images depicting ADEs, aiding healthcare professionals in visually identifying adverse events. Using our MMADE dataset, we showcase the significance of integrating visual cues from images to enhance overall performance. This approach holds promise for patient safety, ADE awareness, and healthcare accessibility, paving the way for further exploration in personalized healthcare. The code and dataset used in this work are publicly available ~\footnote{\url{https://github.com/singhayush27/MMADE.git}}.

\textbf{Disclaimer:} The article features images that may be visually disturbing to some readers.

\end{abstract}

\section{Introduction}
\begin{figure}
  \centering
  \includegraphics[width=\linewidth]{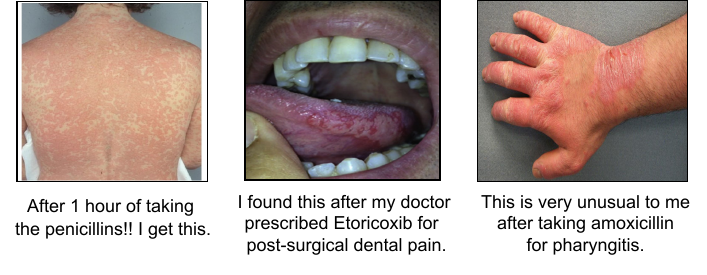}
  \caption{Samples from the dataset highlight the significance of visual cues in understanding adverse drug events, particularly in cases where patients are unaware of a specific medical condition.}
  
  \label{fig:adr1}
\end{figure}

An adverse drug event (ADE) encompasses any harm resulting from medication use, whether it's unintended, off-label, or due to medication errors. Adverse drug reactions (ADRs) are a specific type of ADE, denoting unexpected harm arising from the proper use of medication at the prescribed dosage. Injuries from inappropriate or off-label use are not classified as ADRs~\cite{karimi2015text}. ADEs pose significant public health concerns, contributing to numerous fatalities, serious injuries, millions of hospitalizations, and prolonged hospital stays. Consequently, they impose substantial financial burdens, costing healthcare systems billions of dollars globally. Despite advancements in healthcare, ADE detection remains a significant challenge. Implementing effective detection and monitoring strategies can substantially mitigate the adverse impacts on patients and healthcare systems~\cite{hakkarainen2012percentage},~\cite{yadav2018multi},~\cite{sultana2013clinical}.
Most of the previous ADE detection works are based on text data only~\cite{d2023biodex},~\cite{sarker2015portable},~\cite{sarker2016social},~\cite{chowdhury2018multi},~\cite{yadav2018feature}, which presents a significant disadvantage due to its subjective nature and lack of specific details of visual cues, leading to potential inaccuracies and incomplete categorization of ADE detection. Despite extensive research, the potential of integrating textual data with visual information, such as images, has been largely overlooked. Visual aids are essential in ADE detection for numerous reasons. A substantial proportion of the population lacks proficiency in medical jargon, hindering accurate symptom descriptions. Moreover, certain symptoms are inherently challenging to express through text alone. Patients may struggle to differentiate between similar symptoms, like skin rash, eczema, peeling, and blister. As depicted in Fig.~\ref{fig:adr1}, sample images may present confusion to individuals lacking adequate medical expertise. Integrating both text and images in these scenarios can enhance the accuracy and effectiveness of ADE detection, offering a comprehensive understanding of the patient's current medical condition. To the best of our knowledge, ADE detection using both image and text data has not been explored previously, and we take this opportunity to introduce a MultiModal Adverse Drug Event (MMADE) dataset comprising ADR images paired with corresponding textual descriptions.

Large Language Models (LLMs) and Vision Language Models (VLMs) have exhibited remarkable skills in generating human-like text, prompting their integration into various medical applications, including tasks such as chest radiography report generation, summarization, and medical question answering~\cite{thawkar2023xraygpt},~\cite{ghosh2024medsumm},~\cite{sahoo2024systematic},~\cite{ghosh2023clipsyntel}. However, their potential in ADE detection, which involves both text and images, has yet to be explored. Leveraging LLMs and VLMs for this task presents inherent limitations as they are predominantly trained on generic natural images sourced from databases like ImageNet, Wikipedia, and the internet. Generic models may not possess the specialized medical knowledge required for comprehensive caption generation, potentially leading to oversimplified descriptions that overlook essential details like symptoms and medical intricacies. Furthermore, while VLMs have excelled in traditional visual-linguistic tasks, their application to medical imaging presents unique challenges that may hinder the accurate interpretation and description of complex medical images~\cite{sahoo2024unveiling}. Specialized models such as XrayGPT~\cite{thawkar2023xraygpt} and SkinGPT4~\cite{zhou2023skingpt}, which are trained on chest X-ray and skin disease images, exemplify the domain specificity required for accurate medical image analysis. This has led us to explore ADR detection within a multimodal framework. To support this exploration, we introduce MMADE, a carefully curated dataset crafted for this specific purpose. MMADE consists of 1500 instances of patient-reported concerns regarding drugs and associated side effects, each paired with both textual descriptions and corresponding images. In our study, we have employed InstructBLIP~\cite{dai2023instructblip}, which builds upon the strong foundation of BLIP-2~\cite{li2023blip}, a pre-trained model with high-quality visual representation and strong language generation capabilities. The meticulous fine-tuning process enables it to bridge the disparity between general-purpose models and the specialized demands of ADE-specific tasks. Moreover, our exploration of BLIP~\cite{li2022blip} and GIT~\cite{wang2022git} reveals that these models exhibit insufficient performance before fine-tuning. Nevertheless, upon fine-tuning with domain-specific data, their performance experiences notable improvement.

Our key contributions are as follows:
\begin{itemize}
\item A novel approach to ADE detection in multimodal settings greatly assists medical professionals, such as doctors, nurses, and pharmacists, by delivering detailed descriptions of ADE cases, enhancing precision in diagnosis, treatment planning, and patient care.
    \item Introduction of a novel multimodal dataset MMADE for further research on ADE detection area.
    \item The proposed dataset demonstrates promising potential for various applications, including ADE classification, caption generation, and summarization tasks.
    \item We have utilized InstructBLIP and experimented with two other pre-trained VLMs, and reported a detailed analysis.
    \item The ADE-specific model holds promise for enhancing patient safety, ADE awareness, and healthcare communication. It aims to provide individuals seeking information about ADEs with understandable and informative captions accompanying medical images to improve their comprehension of potential medication risks.
\end{itemize}

\section{Related Works}
This section details the related ADE detection tasks based on the data sources. \\
\textbf{Works Based on Biomedical Text and Electronic Medical Record:}
Various techniques have been developed for extracting ADEs from Electronic Medical Records (EMRs)~\cite{aramaki2010extraction},~\cite{wang2009active}, as well as from medical case reports (MCRs)~\cite{gurulingappa2011identification}.~\citet{gurulingappa2012extraction} utilized machine learning methods to identify and extract potential ADE relations from MEDLINE case reports. Unlike random data sources such as social media, both EMRs and MCRs offer significant advantages by providing comprehensive records of a patient's medical history, treatment, conditions, and potential risk factors. Moreover, these records are not limited to patients who have experienced ADRs~\cite{harpaz2013combing}.~\citet{sarker2015portable} conducted a study by taking data from MEDLINE case reports and Twitter and reported how combining different datasets increases the performance of identifying ADRs.~\citet{huynh2016adverse} explored various neural network frameworks for ADE classification, utilizing datasets from both MCRs and Twitter. DISAE is another corpus~\cite{gurulingappa2010empirical}, which consists of 400 MEDLINE articles with annotations for disease and adverse effect names without drug-related information.~\citet{d2023biodex} introduced the BioDEX dataset for biomedical ADE extraction in real-world pharmacovigilance. This dataset comprises 65,000 abstracts, 19,000 full-text biomedical papers, and 256,000 document-level safety reports crafted by medical professionals. However, to the best of our knowledge, there is currently no publicly accessible annotated multimodal (Image and Text) corpus suitable for identifying drug-related adverse effects.

\textbf{Works Based on Social Media Datasets:}
Social media has become vital for accessing vast amounts of real-time information, making it valuable for identifying potential ADEs.~\citet{leaman2010towards} conducted a pioneering study that analyzed user comments from social media posts, comprising a dataset of 6890 comments. The research demonstrated the significant value of user comments in identifying ADEs, highlighting their crucial role in this context. Several other authors~\cite{gurulingappa2012development},~\cite{yadav2020relation},~\cite{benton2011identifying} employed lexicon-based approaches to extract ADEs. However, these methods are limited to a specific set of target ADEs.~\citet{nikfarjam2011pattern} employed a rule-based technique instead of a naive lexicon-based approach on the same dataset, enabling the detection of ADEs not covered by lexicons. Several authors utilized supervised machine learning techniques like Support Vector Machines (SVM)~\cite{sarker2015portable}, Conditional Random Fields (CRF)~\cite{nikfarjam2015pharmacovigilance}, and Random Forests~\cite{zhang2016ensemble} for ADE detection.~\citet{sarker2016social} introduced one corpus by collecting data from social media, focusing on adverse drug reactions. Tasks included automatically classifying user posts, extracting specific mentions, and normalizing mentions to standardized concepts. With the availability of annotated data, in recent times, the rise of deep learning techniques has significantly influenced research methodologies, leading to the adoption of deep learning models for predicting ADEs.~\citet{tutubalina2017combination} explored the synergy between CRF and Recurrent Neural Networks (RNN), demonstrating that CRF enhances the RNN model's ability to capture contextual information effectively. \citet{chowdhury2018multi} developed a multi-task architecture that simultaneously tackled binary classification, ADR labeling, and indication labeling, using the PSB 2016 Social Media dataset~\cite{sarker2016social}.

 \begin{table}
\begin{center}
\caption{Keywords used for the data collection.}
\label{keyword}
\scalebox{1}{
\begin{tabular}{|c|c|}
\hline
\textbf{Keywords}: ADE, ADR, adverse drug reaction, \\ adverse drug event, adverse reaction,\\
adverse drug event reporting, side effects, \\ drug reactions, drug side effects, 
\\
type of infection and reaction, medicine, \\ drugs, skin rashes, red patches,\\ 
 eczema, ulcer, acne, skin irritation, \\ edema, rosacea, alopecia, lip swelling.\\

\hline
\end{tabular}
}
\end{center}
\end{table}

\section{Corpus Development}
\label{CD}
Our study began with a thorough literature review to identify existing ADE-related datasets. We discovered four text-only datasets: PSB 2016 social media shared task~\cite{sarker2016social} comprising 572 tweets, Medline ADE corpus~\cite{gurulingappa2012development} with 4,272 sentences, CADEC~\cite{karimi2015cadec} containing 1,248 sentences, and recently released BioDEX dataset~\cite{d2023biodex}. This revealed a notable gap in multimodal ADE datasets, where images complement textual data. We take this opportunity to introduce a multimodal corpus consisting of 1,500 ADE images with corresponding English sentences to facilitate further research. While preparing this corpus, we carry out the following steps.




\subsection{Data Collection}
Utilizing a diverse array of keywords, we have curated a comprehensive dataset from social media, healthcare blogs, and MCRs (refer to Table~\ref{keyword}). The inclusion of various sources ensures a broad representation of the population, enriching the dataset's diversity. 


\begin{figure}
  \centering
  \includegraphics[width=\linewidth]{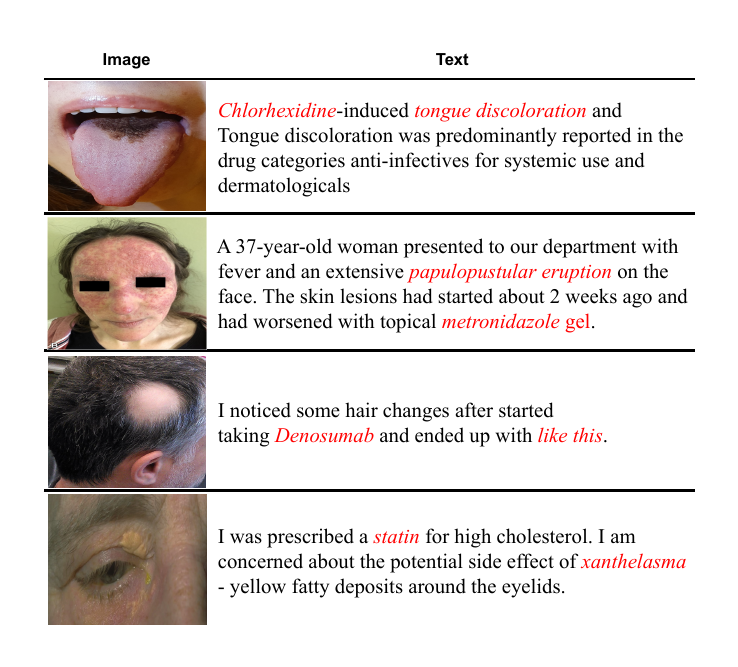}
  \caption{Some samples from the dataset containing images and the corresponding descriptive text.}
  \label{fig:archi}
\end{figure}

\subsubsection{Social Media}
Social media data is invaluable for ADE-related tasks due to its real-time nature and diverse user-generated content~\cite{sarker2016social}. In this study, we have utilized the official X (Twitter) and scraper API to gather Tweets\footnote{\url{https://twitter.com/}},\footnote{\url{https://www.scraperapi.com/}} employing diverse keywords related to ADE. The data collection phase, conducted between June 2023 and October 2023, collected a total of 20,000 tweets using specified keywords presented in Table~\ref{keyword}. From this pool, 3,000 tweets were meticulously identified as pertinent to ADEs, featuring either images, text, or a combination of both. Notably, 142 tweets included relevant images accompanied by textual descriptions of the adverse drug events.






\subsubsection{Healthcare Blog}
We utilized a public healthcare-related blog, healthdirect\footnote{\url{https://www.healthdirect.gov.au/}}, a government-funded virtual health service that provides access to health advice and information via a website. We used Python's BeautifulSoup library to scrape the data and collected 1,150 unique images with corresponding text. Among these, 54 relevant images depicting adverse drug events were manually curated along with corresponding texts.



\begin{figure*}
  \centering
  \includegraphics[width=0.90\linewidth]{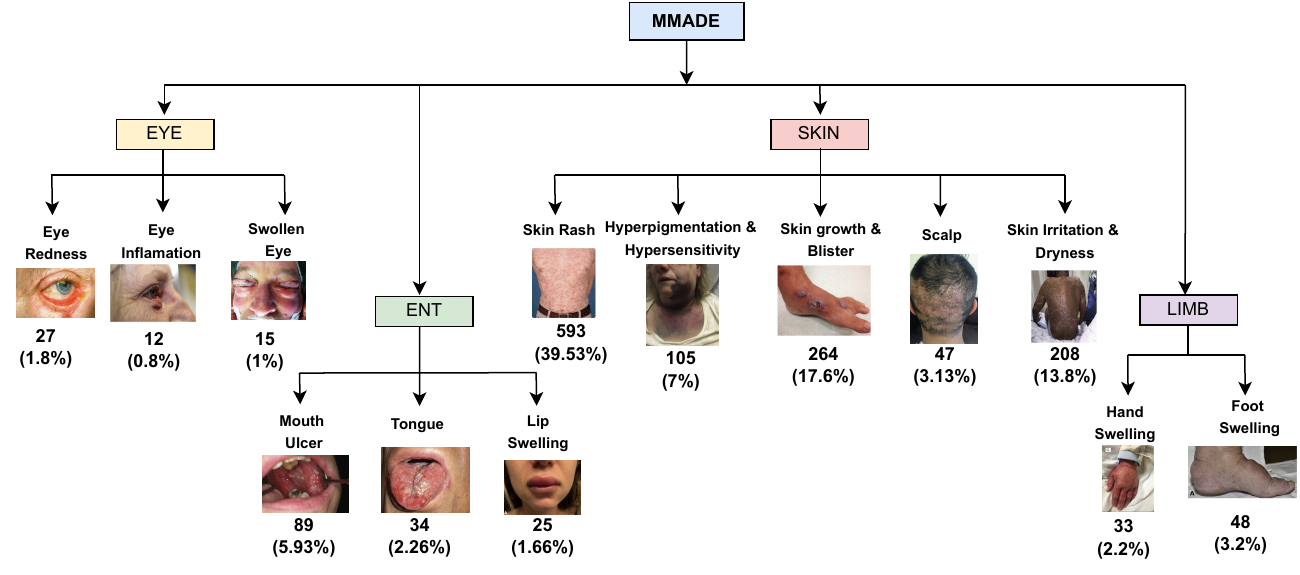}
  \caption{Distribution of different body parts in the curated MMADE dataset. The number of data points and the percentage corresponding to each category have been provided.}
  \label{fig:MMADE}
\end{figure*}

\subsubsection{Medical Case Reports}
Data from published medical case reports is crucial for constructing comprehensive datasets to analyze ADE. These articles provide structured and verified information, forming a reliable foundation for in-depth analysis and research in pharmacovigilance. We have extracted data from the New England Journal of Medicine\footnote{\url{https://www.nejm.org}}, Science Direct\footnote{\url{https://www.sciencedirect.com}}. 
A precise Science Direct query performed is as follows: \\

\texttt{(("adverse drug event") AND (Languages=English) AND (Article \\type=Case Reports) AND (Years=2000 to 2023))}
\\

This approach retrieved approximately 2,907 documents from ScienceDirect, from which we manually selected 1390 relevant images with corresponding texts.

\begin{figure}
  \centering
 \scalebox{1} {\includegraphics[width=\linewidth]{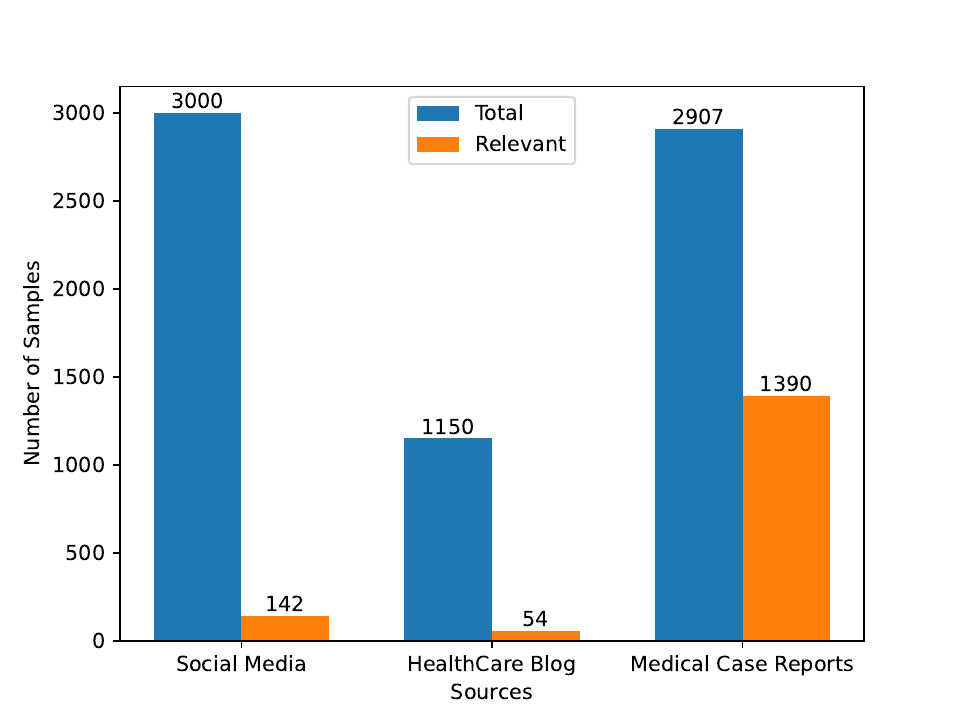}}
  \caption{Data distribution statistics from various sources, illustrating the total ADE data (images, text, and image-text pairs), and the relevant image-text pairs.}
  \label{fig:dis}
\end{figure}


\subsection{Data Annotation} 
To ensure meticulous annotation aligned with ethical standards, we enlisted the assistance of two medical students and one Ph.D. student, selected based on specific criteria. These criteria included being at least 25 years old, proficient in English (reading, writing, and speaking), and willing to handle sensitive content. The process was finalized within a span of five months, and participants received compensation for their involvement~\footnote{The medical students received compensation in the form of gift vouchers and honorarium amounts in accordance with~\url{https://www.minimum-wage.org/international/india.}}. 
Annotators were tasked with meticulously assessing each image and its corresponding text based on the annotation manual. Sentences accurately depicting adverse drug events, including the drug's name and associated side effects, were chosen for inclusion in the corpus development process, while all other instances were deemed irrelevant and removed. To maintain consistency and consensus among annotators, final rationale labels were determined through a majority voting approach. Annotators were explicitly instructed to annotate posts without biases related to demographics, religion, or other extraneous factors. The quality of annotations was assessed by measuring inter-annotator agreement (IAA) using Cohen's Kappa score~\cite{viera2005understanding}. The resulting agreement scores of 0.78 affirm the acceptability and high quality of the annotations.

\subsection{Annotation Manual}
\label{an}
\begin{figure}
  \centering
  \includegraphics[width=1\linewidth]{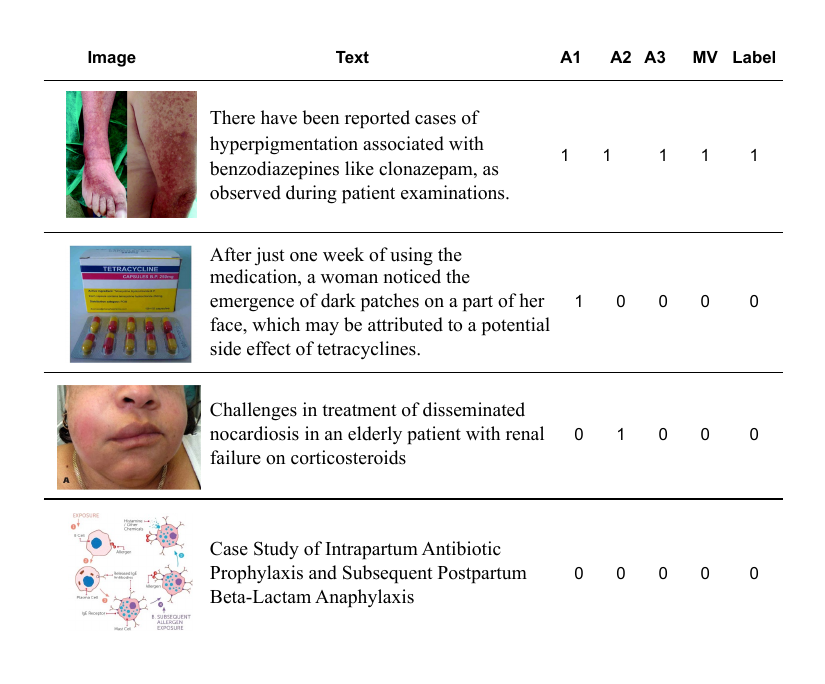}
  \caption{Illustration of an annotation process, with A1, A2, and A3 representing individual annotators and MV signifying final majority voting. In this setup, "1" denotes ADR, while "0" is unrelated to ADR.}
  \label{annotation1}
\end{figure}

We initially provided annotators with an instruction manual detailing different instances, accompanied by examples (as depicted in Fig.~\ref{annotation1}).
\begin{itemize}
    \item For each data instance comprising text and an image, select the data instance if both the text and image indicate concerns about drug side effects.
    \item Remove data instances where the image does not convey the side effects mentioned in the text.
    \item Remove data instances where the text does not convey any side effects related to drugs, but some side effects are visually present in the accompanying image.
    \item  Remove instances where both the text and image are unrelated to any drug side effect.
    \item If a data instance includes a URL link, access the content at that URL address to gain additional insight and context about the data.
\end{itemize}

\label{co}

\subsection{Corpus Analysis}
\textls[-25]{Figure~\ref{fig:dis} provides an overview of the data sources and the distribution of relevant cases within the dataset. Following the initial collection of 7,057 ADE-related samples encompassing images, text, and image-text pairs, we curated 1,500 pertinent samples. These encompass both image and their associated text descriptions of ADEs, establishing the foundation of our multimodal ADE dataset. Figure~\ref{fig:archi} depicts some of the samples from the dataset. The first sample shows a person suffering from tongue discoloration after taking a Chlorhexidine drug. Both the text and visual image show a clear indication of drug reaction. The second sample is of a woman with papulopustular eruption on the face, which had worsened with topical metronidazole gel. Adverse drug events can manifest internally or externally, yet acquiring images of internal body parts from public sources is challenging. Thus, we focus on external body parts or symptoms, which can be readily captured and shared with doctors, pharmacovigilance teams, or the public domain for improved consultation and advice. Following dataset analysis, we identified 13 significant adverse effects crucial for multimodal ADE reporting. These effects are categorized into four groups based on their origin: ENT (9.85\%), EYE (3.6\%), LIMB (5.4\%), and SKIN (81.06\%). Additional details of the dataset, such as the distribution of samples across different body parts along with corresponding percentages, are illustrated in Fig.~\ref{fig:MMADE}.}

\begin{figure*}
  \centering
  \includegraphics[width=\linewidth]{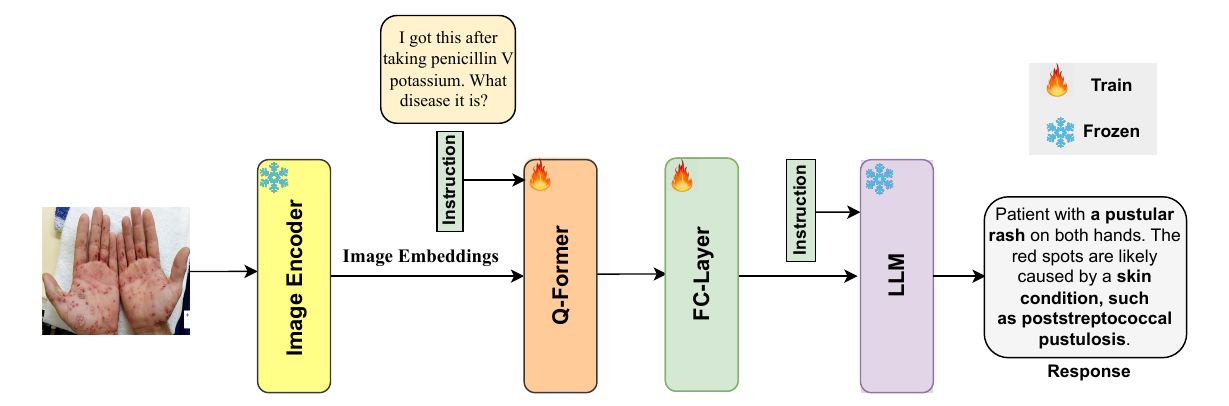}
  \caption{The architectural details are illustrated, depicting the input image and corresponding text. Additionally, the frozen and trainable layers are highlighted for clarity.}
  \label{archii}
\end{figure*}

\section{Problem Formulation} Each data point in the dataset encompasses a patient's textual description $T$ along with a corresponding image $I$ illustrating their medical issue or concern visually. The textual representation comprises a sequence of words $\{t_1, t_2, \ldots, t_n\}$, while the visual elements are represented by $I \in \mathbb{R}^{3 \times W \times H}$, where $W$ and $H$ denote the width and height of the image data, respectively. The objective is to process both the text $T$ and image $I$ for each patient and generate a natural language sequence $Y$ that seamlessly integrates both modalities, expressed as $Y = \{T, I\}$. 

\section{Methodology}
\label{me}
Recent advancements in VLMs, such as BLIP, InstructBlip, and GIT, have showcased remarkable advancements in encoding both textual and visual inputs. These models outperform traditional approaches that rely solely on individual image or text encoders, followed by fusion. By integrating sophisticated mechanisms for joint representation learning, VLMs excel at capturing intricate relationships between textual and visual modalities, thereby enhancing their ability to generate more contextually relevant and coherent outputs~\cite{zhang2023visionlanguage}. In the proposed work, we have leveraged InstructBlip~\cite{dai2023instructblip}, known for its exceptional performance across a range of vision-language (VL) tasks including Visual Question Answering (VQA), Image captioning, and Image retrieval. Each patient's inquiry or concern is articulated as a textual sentence along with a visual image where they elaborate on their medical queries or concerns on social media platforms or healthcare blogs to obtain pertinent feedback or advice. InstructBlip integrates two distinct encoders specialized for separate modalities. This architecture includes an image encoder, which utilizes a vision transformer (ViT) to extract visual features, alongside an LLM and a Query Transformer (Q-Former). The Q-Former interacts with the image encoder's output through cross-attention, resulting in K-encoded visual vectors, which are then linearly projected and fed into the frozen LLM for further processing. This representation serves as input for a proficient language model, which generates high-quality text. During instruction tuning, only the Q-Former undergoes fine-tuning, while the image encoder and LLM remain unchanged. Figure~\ref{archii} illustrates the detailed architecture. We fine-tuned the InstructBLIP to assess its performance on the proposed MMADE dataset. InstructBLIP maintains a consistent image resolution $(224\times224)$ during instruction tuning and freezes the visual encoder during fine-tuning. This approach substantially reduces the number of trainable parameters from 1.2 billion to 188 million, improving fine-tuning efficiency. We also employed two more VLMs and fine-tuned them with the proposed MMADE dataset. BLIP~\cite{li2022blip}, a versatile vision-language model, utilizes a multimodal mixture of encoder-decoder models during pre-training. This involves bootstrapping a dataset from large-scale noisy image-text pairs, where synthetic captions are injected, and noisy captions are eliminated. Additionally, we utilized GIT~\cite{wang2022git}, a VLM model that generates text descriptions of images. Trained on curated datasets of image-text pairs, GIT encodes image features into a latent representation, which is decoded into text descriptions. Its architecture comprises an image encoder using a pre-trained Swin transformer and a text decoder based on a standard transformer decoder, linked by a cross-attention layer for enhanced focus on specific image encoding parts. Additionally, we performed the integration of LSTM networks with VGG16~\cite{simonyan2014very} and ResNet50~\cite{he2016deep} architectures. In this setup, the LSTM serves as the text encoder, while either VGG16 or ResNet50 acts as the visual encoder. The features extracted from the text and visual encoders are then concatenated to create a joint representation, allowing for comprehensive modeling of both textual and visual information. Please refer to Section~\ref{FT} of the Appendix for the fine-tuning details.

\section{Experimental Results and Analysis}
\label{ex}
We have utilized four commonly employed evaluation metrics to assess the performance of the models, including BLEU score (Bilingual Evaluation Understudy)~\cite{papineni2002bleu}, ROUGE score (Recall-Oriented Understudy for Gisting Evaluation)~\cite{lin2004rouge}, BERTScore~\cite{zhang2019bertscore}, and MoverScore~\cite{zhao2019moverscore}. Detailed explanations of experimental settings and metrics are given in Section~\ref{es} and~\ref{em} of the Appendix.

\begin{table}
\begin{center}
\caption{The performance evaluation of the models using ROUGE and BLEU Scores in the multimodal dataset setting.}
\label{rouge}
\scalebox{0.58}{
\begin{tabular}{cccccccc}
\toprule
\multicolumn{1}{c}{\multirow{2}{*}{\textbf{Model}}}&
\multicolumn{1}{c}{\multirow{2}{*}{\textbf{Type}}}&
\multicolumn{3}{c}{\textbf{ROUGE}} &
\multicolumn{3}{c}{\textbf{BLEU}} 
\\
\cmidrule{3-8}
    &&\textbf{R1} &\textbf{R2} & \textbf{RL} &\textbf{B1} &\textbf{B2} &\textbf{B3} \\
\midrule
LSTM+VGG16&-& 0.213 & 0.105 & 0.201 & 0.165 & 0.073 & 0.041 \\
LSTM+ResNet50&-&0.281 & 0.086 & 0.230 & 0.179 & 0.058 &0.046 \\
BLIP& Base & 0.19& 0.093 &0.185 & 0.099&0.003&0.001\\\
& Fine-Tune & 0.334 & 0.163 & 0.225 &0.171 & 0.081 &0.058  \\
GIT& Base & 0.27& 0.11 &0.192 &0.157&0.014  & 0.004\\
& Fine-Tune & 0.504 & 0.285 & 0.416 &0.194&0.10&0.097\\
InstructBLIP& Base & 0.29& 0.161 &0.212 & 0.219&0.125&0.008\\
& Fine-Tune & \textbf{0.571} & \textbf{0.351}& \textbf{0.475} & \textbf{0.319}&\textbf{0.175} & \textbf{0.112} \\
\bottomrule

\end{tabular}
}
\end{center}
\end{table}

\begin{table}
\begin{center}
\caption{The evaluation of the model using BERTScore and MoverScore in the multimodal dataset setting.}
\label{bert}
\scalebox{0.45}{
\begin{tabular}{ccccccccc}
\toprule
\multicolumn{9}{c}{\textbf{BERTScore}} \\
\cmidrule{1-9}

\multicolumn{1}{c}{\multirow{2}{*}{\textbf{Evaluation Metrics}}}&
\multicolumn{1}{c}{\multirow{1}{*}{\textbf{\makecell{LSTM+\\VGG16}}}}&
\multicolumn{1}{c}{\multirow{1}{*}{\textbf{\makecell{LSTM+\\ResNet50}}}}&
\multicolumn{2}{c}{\textbf{BLIP}} &
\multicolumn{2}{c}{\textbf{GIT}} &
\multicolumn{2}{c}{\textbf{InstructBLIP}} \\
\cmidrule{4-9}
    & & &\textbf{Base} & \textbf{Fine-Tune} & \textbf{Base} & \textbf{Fine-Tune} & \textbf{Base} &\textbf{Fine-Tune} \\
\midrule
Precision &0.821&0.847& 0.819 & 0.841 & 0.826 & 0.866 & 0.832 & \textbf{0.896}\\
Recall &0.797&0.811 & 0.783 & 0.819 & 0.791 & 0.831 & 0.781 & \textbf{0.891} \\
F1-score &0.809&0.821& 0.801 & 0.829 & 0.812 & 0.852 & 0.805 & \textbf{0.893} \\

\midrule
\multicolumn{9}{c}{\textbf{MoverScore}} \\
\cmidrule{1-9}
 &0.441&0.487& 0.482 & 0.513 & 0.496 & 0.553 & 0.544 & \textbf{0.622} \\
\bottomrule
\end{tabular}
}
\end{center}
\end{table}

\begin{table}
\begin{center}
\caption{The evaluation of the models using BERTScore and MoverScore in the unimodal dataset setting.}
\label{bertuni}
\scalebox{0.5}{
\begin{tabular}{cccccc}
\toprule
\multicolumn{6}{c}{\textbf{BERTScore}} \\
\cmidrule{1-6}
\textbf{Evaluation Metrics}&
\textbf{LSTM+VGG16}&
\textbf{LSTM+ResNet50}&
\multicolumn{1}{c}{\textbf{BLIP}} &
\multicolumn{1}{c}{\textbf{GIT}} &
\multicolumn{1}{c}{\textbf{InstructBLIP}} \\
Precision & 0.731 &0.744 & 0.753 & 0.801 & 0.840  \\
Recall & 0.709 & 0.713 & 0.724 & 0.783 & 0.783 \\
F1-score & 0.719 &0.729 & 0.738 & 0.790 & 0.810  \\
\bottomrule
\multicolumn{6}{c}{\textbf{MoverScore}} \\
\cmidrule{1-6}
 & 0.121&0.143 & 0.151 & 0.184 & 0.253  \\
\bottomrule
\end{tabular}
}
\end{center}
\end{table}

\subsection{Findings from Experiments}
\label{result}
The experimental results across Tables~\ref{rouge} and~\ref{bert} depict the performance of various models in multimodal dataset settings, while Tables~\ref{bertuni} and~\ref{ablationtable} represent the performance in unimodal dataset settings, based on standard evaluation metrics. From the results, several observations are:




\begin{itemize}
    \item Table~\ref{rouge} presents the ROUGE and BLEU scores, indicating the superior performance of fine-tuned InstructBLIP compared to other models. The higher ROUGE and BLEU scores achieved by fine-tuned InstructBLIP suggest its proficiency in capturing relevant information from the input and generating text. This superiority can be attributed to its training on more extensive and diverse datasets, enabling it to capture intricate data patterns effectively.
    \item Tables~\ref{bert} and~\ref{bertuni} present BERTScores and MoverScores in the multimodal and unimodal settings, respectively. Fine-tuned InstructBLIP achieves the highest BERTScore and MoverScores, demonstrating its efficacy in capturing contextual similarity and superior ability to convey meaningful content effectively.
    
    



    
    \item  Table~\ref{bertuni} and~\ref{ablationtable} show the decline in performance when training VLM models with unimodal data. The absence of meaningful visual information likely contributes to the performance degradation. 

    \item One important observation is that InstructBLIP consistently outperforms BLIP and GIT across all metrics, showing its superior capability in effectively integrating textual and visual information. This can be attributed to its innovative architecture, featuring a Query Transformer for instruction-aware feature extraction. Refer to Section~\ref{co} of the Appendix for detailed outcomes.
    
    \item Fine-tuning with domain-specific ADE data remarkably enhances model performance, reflecting its pivotal role in adapting models to the intricacies of adverse drug event detection.

    \item Another key finding highlights the substantial performance enhancement achieved by integrating both image and text modalities, emphasizing the critical role of visual information alongside textual data. 
    




\end{itemize}

{\em Statistical Analysis:} We conducted a statistical t-test to compare the performance of the proposed multimodal model, utilizing both image and text data, with that of unimodal models. The analysis yielded a p-value below 0.05, indicating a significant difference in performance between the two. The detailed explanation is given in Section~\ref{SA} of Appendix.

\begin{figure}
  \centering
  \includegraphics[width=\linewidth]{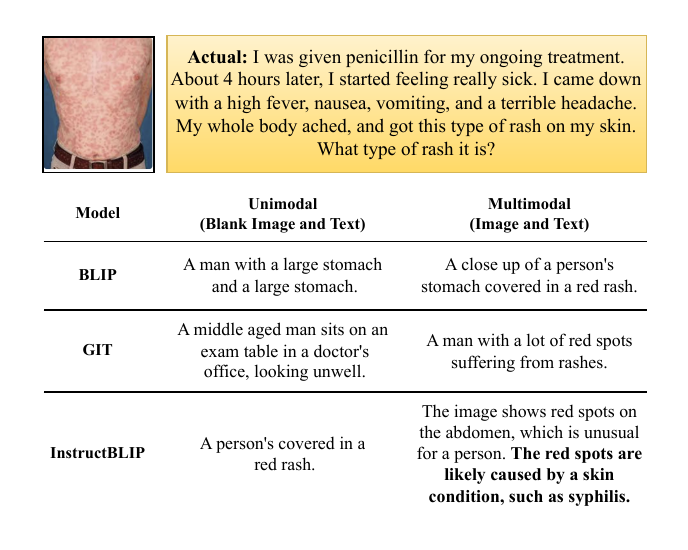}
  \caption{An example case study where a sample image and its corresponding text are provided along with the generated text from all models in both unimodal and multimodal settings.}
  \label{ablation}
\end{figure}

\subsection{Qualitative Analysis} 
\label{QA}
We performed an extensive qualitative analysis of the responses generated by various models in both unimodal and multimodal settings, complemented by several case studies. A detailed case study is depicted in Fig.~\ref{ablation}. The analysis has led to the following conclusion: (a) In multimodal settings, all the models demonstrate superior performance and exhibit a greater ability to capture crucial visual information conveyed through images than in unimodal settings (refer to Table~\ref{bertuni} and~\ref{ablationtable}).  (b) Observations also revealed that models such as BLIP and GIT tended to hallucinate, as evidenced by Fig.~\ref{ablation}, occasionally generating facts that were entirely unrelated to the context. (c) InstructBLIP demonstrates superior performance, leading us to conclude that providing instructions during fine-tuning prompts the model to selectively focus on pertinent visual features. This focused attention encourages the model to generate target sequences that closely resemble the desired output. 
We have presented a case study demonstrating that model performance is not solely determined by the number of samples in the dataset but also by other factors such as the distinct visual characteristics of different types of ADEs present in the dataset. A detailed explanation is provided in the Section~\ref{cs} of Appendix.


\section{Risk Analysis}
While our multimodal model demonstrates promise, it is essential to have medical experts and pharmacovigilance teams validate the findings, considering other critical factors. Our model and dataset are intended to support medical professionals rather than replace them.

\begin{table}
\begin{center}
\caption{The performance evaluation of the models
using ROUGE and BLEU Scores in the unimodal
dataset setting.}
\label{ablationtable}
\scalebox{0.7}{
\begin{tabular}{ccccccc}
\toprule
\multicolumn{1}{c}{\multirow{2}{*}{\textbf{Model}}}&
\multicolumn{3}{c}{\textbf{ROUGE}} &
\multicolumn{3}{c}{\textbf{BLEU}} 
 \\
\cmidrule{2-7}
    &\textbf{R1} &\textbf{R2} & \textbf{RL} &\textbf{B1} &\textbf{B2} &\textbf{B3} \\
\midrule
LSTM+VGG16&0.133 & 0.018 & 0.101 & 0.141& 0.012 &0.001  \\
LSTM+ResNet50&0.146&0.029&0.112&0.152 & 0.016&0.002\\
BLIP& 0.158& 0.032 &0.129 &0.161&0.019 & 0.002\\
GIT&0.169 &0.035 &0.137 & 0.172&0.020&0.003\\
InstructBLIP& 0.211& 0.087 & 0.172&0.197 &0.089 &0.006\\
\bottomrule

\end{tabular}
}
\end{center}
\end{table}


\section{Conclusion and Future Work}
\label{cf}
In this paper, we present the task of ADE detection within pharmacovigilance mining, leveraging multimodal datasets. In order to solve this task, we have created a multimodal ADE dataset, MMADE, containing images and corresponding descriptions, enhancing decision-making with the inclusion of visual cues. We have employed InstructBLIP, fine-tuned with the proposed dataset, and compared it with other models. Our findings suggest that domain-specific fine-tuning significantly enhances overall performance, emphasizing the importance of multimodal visual cues. We envision MMADE as a pivotal resource for advancing research in multimodal ADE detection. Moreover, our fine-tuned architecture holds promise as a valuable tool for pharmacovigilance teams, clinicians, and researchers, facilitating more effective ADE monitoring and ultimately improving patient safety and outcomes. In addition to expanding the dataset, future investigations could explore the potential of this multimodal dataset in tasks such as ADE severity classification and summarization.

\section{Limitations}
\label{sec:limitation}
While our effort aimed to develop an ADE detection framework and introduce the novel MMADE dataset, comprising textual descriptions of drug events paired with images, it is crucial to acknowledge certain limitations inherent in the dataset. Specifically, our dataset primarily focuses on drug events associated with external body parts, omitting data about internal conditions such as liver infections, kidney stones, or psychological ailments like depression and migraine. Acquiring a substantial volume of image-text pairs within the ADE domain presents inherent challenges, including data privacy concerns, regulatory constraints, and the specialized nature of ADE occurrences. Despite these obstacles, our research breaks new ground by integrating images with text, solving a real-world challenge where individuals affected by ADE may resort to image sharing for communication when verbally expressing their symptoms is difficult. Moving forward, we aim to enhance the dataset by incorporating more ADE-related images and expanding its utility through additional tasks such as complaint identification and text summarization.

\section{Ethics and Broader Impact}

\subsubsection*{User Privacy.}
Our dataset contains AD images and corresponding text tweets with annotation labels and no personal user information.

\subsubsection*{Biases.}
Any biases detected in the dataset are inadvertent, and we have no intention of harming anyone or any group. We acknowledge that evaluating whether a tweet is ADE can be subjective, so we have taken agreement from all the annotators before selecting the data.


\subsubsection*{Intended Use.}
We share our data to promote more research on Adverse drug event detection. We only release the dataset for research purposes and do not grant a license for commercial use.

\bibliography{custom}

\begin{thebibliography}{44}
\expandafter\ifx\csname natexlab\endcsname\relax\def\natexlab#1{#1}\fi

\bibitem[{Aramaki et~al.(2010)Aramaki, Miura, Tonoike, Ohkuma, Masuichi, Waki, and Ohe}]{aramaki2010extraction}
Eiji Aramaki, Yasuhide Miura, Masatsugu Tonoike, Tomoko Ohkuma, Hiroshi Masuichi, Kayo Waki, and Kazuhiko Ohe. 2010.
\newblock Extraction of adverse drug effects from clinical records.
\newblock In \emph{MEDINFO 2010}, pages 739--743. IOS Press.

\bibitem[{Benton et~al.(2011)Benton, Ungar, Hill, Hennessy, Mao, Chung, Leonard, and Holmes}]{benton2011identifying}
Adrian Benton, Lyle Ungar, Shawndra Hill, Sean Hennessy, Jun Mao, Annie Chung, Charles~E Leonard, and John~H Holmes. 2011.
\newblock Identifying potential adverse effects using the web: A new approach to medical hypothesis generation.
\newblock \emph{Journal of biomedical informatics}, 44(6):989--996.

\bibitem[{Chowdhury et~al.(2018)Chowdhury, Zhang, and Yu}]{chowdhury2018multi}
Shaika Chowdhury, Chenwei Zhang, and Philip~S Yu. 2018.
\newblock Multi-task pharmacovigilance mining from social media posts.
\newblock In \emph{Proceedings of the 2018 World Wide Web Conference}, pages 117--126.

\bibitem[{Dai et~al.(2023)Dai, Li, Li, Tiong, Zhao, Wang, Li, Fung, and Hoi}]{dai2023instructblip}
Wenliang Dai, Junnan Li, Dongxu Li, Anthony Meng~Huat Tiong, Junqi Zhao, Weisheng Wang, Boyang Li, Pascale Fung, and Steven Hoi. 2023.
\newblock \href {http://arxiv.org/abs/2305.06500} {Instructblip: Towards general-purpose vision-language models with instruction tuning}.

\bibitem[{D'Oosterlinck et~al.(2023)D'Oosterlinck, Remy, Deleu, Demeester, Develder, Zaporojets, Ghodsi, Ellershaw, Collins, and Potts}]{d2023biodex}
Karel D'Oosterlinck, Fran{\c{c}}ois Remy, Johannes Deleu, Thomas Demeester, Chris Develder, Klim Zaporojets, Aneiss Ghodsi, Simon Ellershaw, Jack Collins, and Christopher Potts. 2023.
\newblock Biodex: Large-scale biomedical adverse drug event extraction for real-world pharmacovigilance.
\newblock \emph{arXiv preprint arXiv:2305.13395}.

\bibitem[{Ghosh et~al.(2024{\natexlab{a}})Ghosh, Acharya, Jain, Saha, Chadha, and Sinha}]{ghosh2023clipsyntel}
Akash Ghosh, Arkadeep Acharya, Raghav Jain, Sriparna Saha, Aman Chadha, and Setu Sinha. 2024{\natexlab{a}}.
\newblock \href {https://doi.org/10.1609/AAAI.V38I20.30206} {Clipsyntel: {CLIP} and {LLM} synergy for multimodal question summarization in healthcare}.
\newblock In \emph{Thirty-Eighth {AAAI} Conference on Artificial Intelligence, {AAAI} 2024, Thirty-Sixth Conference on Innovative Applications of Artificial Intelligence, {IAAI} 2024, Fourteenth Symposium on Educational Advances in Artificial Intelligence, {EAAI} 2014, February 20-27, 2024, Vancouver, Canada}, pages 22031--22039. {AAAI} Press.

\bibitem[{Ghosh et~al.(2024{\natexlab{b}})Ghosh, Acharya, Jha, Saha, Gaudgaul, Majumdar, Chadha, Jain, Sinha, and Agarwal}]{ghosh2024medsumm}
Akash Ghosh, Arkadeep Acharya, Prince Jha, Sriparna Saha, Aniket Gaudgaul, Rajdeep Majumdar, Aman Chadha, Raghav Jain, Setu Sinha, and Shivani Agarwal. 2024{\natexlab{b}}.
\newblock \href {https://doi.org/10.1007/978-3-031-56069-9\_8} {Medsumm: {A} multimodal approach to summarizing code-mixed hindi-english clinical queries}.
\newblock In \emph{Advances in Information Retrieval - 46th European Conference on Information Retrieval, {ECIR} 2024, Glasgow, UK, March 24-28, 2024, Proceedings, Part {V}}, volume 14612 of \emph{Lecture Notes in Computer Science}, pages 106--120. Springer.

\bibitem[{Gurulingappa et~al.(2011)Gurulingappa, Fluck, Hofmann-Apitius, and Toldo}]{gurulingappa2011identification}
Harsha Gurulingappa, Juliane Fluck, Martin Hofmann-Apitius, and Luca Toldo. 2011.
\newblock Identification of adverse drug event assertive sentences in medical case reports.
\newblock In \emph{First international workshop on knowledge discovery and health care management (KD-HCM), European conference on machine learning and principles and practice of knowledge discovery in databases (ECML PKDD)}, pages 16--27.

\bibitem[{Gurulingappa et~al.(2010)Gurulingappa, Klinger, Hofmann-Apitius, and Fluck}]{gurulingappa2010empirical}
Harsha Gurulingappa, Roman Klinger, Martin Hofmann-Apitius, and Juliane Fluck. 2010.
\newblock An empirical evaluation of resources for the identification of diseases and adverse effects in biomedical literature.
\newblock In \emph{2nd Workshop on Building and evaluating resources for biomedical text mining (7th edition of the Language Resources and Evaluation Conference)}, pages 15--22.

\bibitem[{Gurulingappa et~al.(2012{\natexlab{a}})Gurulingappa, Mateen-Rajpu, and Toldo}]{gurulingappa2012extraction}
Harsha Gurulingappa, Abdul Mateen-Rajpu, and Luca Toldo. 2012{\natexlab{a}}.
\newblock Extraction of potential adverse drug events from medical case reports.
\newblock \emph{Journal of biomedical semantics}, 3(1):1--10.

\bibitem[{Gurulingappa et~al.(2012{\natexlab{b}})Gurulingappa, Rajput, Roberts, Fluck, Hofmann-Apitius, and Toldo}]{gurulingappa2012development}
Harsha Gurulingappa, Abdul~Mateen Rajput, Angus Roberts, Juliane Fluck, Martin Hofmann-Apitius, and Luca Toldo. 2012{\natexlab{b}}.
\newblock Development of a benchmark corpus to support the automatic extraction of drug-related adverse effects from medical case reports.
\newblock \emph{Journal of biomedical informatics}, 45(5):885--892.

\bibitem[{Hakkarainen et~al.(2012)Hakkarainen, Hedna, Petzold, and H{\"a}gg}]{hakkarainen2012percentage}
Katja~M Hakkarainen, Khadidja Hedna, Max Petzold, and Staffan H{\"a}gg. 2012.
\newblock Percentage of patients with preventable adverse drug reactions and preventability of adverse drug reactions--a meta-analysis.
\newblock \emph{PloS one}, 7(3):e33236.

\bibitem[{Harpaz et~al.(2013)Harpaz, Vilar, DuMouchel, Salmasian, Haerian, Shah, Chase, and Friedman}]{harpaz2013combing}
Rave Harpaz, Santiago Vilar, William DuMouchel, Hojjat Salmasian, Krystl Haerian, Nigam~H Shah, Herbert~S Chase, and Carol Friedman. 2013.
\newblock Combing signals from spontaneous reports and electronic health records for detection of adverse drug reactions.
\newblock \emph{Journal of the American Medical Informatics Association}, 20(3):413--419.

\bibitem[{He et~al.(2016)He, Zhang, Ren, and Sun}]{he2016deep}
Kaiming He, Xiangyu Zhang, Shaoqing Ren, and Jian Sun. 2016.
\newblock Deep residual learning for image recognition.
\newblock In \emph{Proceedings of the IEEE conference on computer vision and pattern recognition}, pages 770--778.

\bibitem[{Huynh et~al.(2016)Huynh, He, Willis, and R{\"u}ger}]{huynh2016adverse}
Trung Huynh, Yulan He, Alistair Willis, and Stefan R{\"u}ger. 2016.
\newblock Adverse drug reaction classification with deep neural networks.
\newblock Coling.

\bibitem[{Karimi et~al.(2015{\natexlab{a}})Karimi, Metke-Jimenez, Kemp, and Wang}]{karimi2015cadec}
Sarvnaz Karimi, Alejandro Metke-Jimenez, Madonna Kemp, and Chen Wang. 2015{\natexlab{a}}.
\newblock Cadec: A corpus of adverse drug event annotations.
\newblock \emph{Journal of biomedical informatics}, 55:73--81.

\bibitem[{Karimi et~al.(2015{\natexlab{b}})Karimi, Wang, Metke-Jimenez, Gaire, and Paris}]{karimi2015text}
Sarvnaz Karimi, Chen Wang, Alejandro Metke-Jimenez, Raj Gaire, and Cecile Paris. 2015{\natexlab{b}}.
\newblock Text and data mining techniques in adverse drug reaction detection.
\newblock \emph{ACM Computing Surveys (CSUR)}, 47(4):1--39.

\bibitem[{Leaman et~al.(2010)Leaman, Wojtulewicz, Sullivan, Skariah, Yang, and Gonzalez}]{leaman2010towards}
Robert Leaman, Laura Wojtulewicz, Ryan Sullivan, Annie Skariah, Jian Yang, and Graciela Gonzalez. 2010.
\newblock Towards internet-age pharmacovigilance: extracting adverse drug reactions from user posts in health-related social networks.
\newblock In \emph{Proceedings of the 2010 workshop on biomedical natural language processing}, pages 117--125.

\bibitem[{Li et~al.(2023)Li, Li, Savarese, and Hoi}]{li2023blip}
Junnan Li, Dongxu Li, Silvio Savarese, and Steven Hoi. 2023.
\newblock Blip-2: Bootstrapping language-image pre-training with frozen image encoders and large language models.
\newblock \emph{arXiv preprint arXiv:2301.12597}.

\bibitem[{Li et~al.(2022)Li, Li, Xiong, and Hoi}]{li2022blip}
Junnan Li, Dongxu Li, Caiming Xiong, and Steven Hoi. 2022.
\newblock Blip: Bootstrapping language-image pre-training for unified vision-language understanding and generation.
\newblock In \emph{International Conference on Machine Learning}, pages 12888--12900. PMLR.

\bibitem[{Lin(2004)}]{lin2004rouge}
Chin-Yew Lin. 2004.
\newblock Rouge: A package for automatic evaluation of summaries.
\newblock In \emph{Text summarization branches out}, pages 74--81.

\bibitem[{Nikfarjam and Gonzalez(2011)}]{nikfarjam2011pattern}
Azadeh Nikfarjam and Graciela~H Gonzalez. 2011.
\newblock Pattern mining for extraction of mentions of adverse drug reactions from user comments.
\newblock In \emph{AMIA annual symposium proceedings}, volume 2011, page 1019. American Medical Informatics Association.

\bibitem[{Nikfarjam et~al.(2015)Nikfarjam, Sarker, O’connor, Ginn, and Gonzalez}]{nikfarjam2015pharmacovigilance}
Azadeh Nikfarjam, Abeed Sarker, Karen O’connor, Rachel Ginn, and Graciela Gonzalez. 2015.
\newblock Pharmacovigilance from social media: mining adverse drug reaction mentions using sequence labeling with word embedding cluster features.
\newblock \emph{Journal of the American Medical Informatics Association}, 22(3):671--681.

\bibitem[{Papineni et~al.(2002)Papineni, Roukos, Ward, and Zhu}]{papineni2002bleu}
Kishore Papineni, Salim Roukos, Todd Ward, and Wei-Jing Zhu. 2002.
\newblock Bleu: a method for automatic evaluation of machine translation.
\newblock In \emph{Proceedings of the 40th annual meeting of the Association for Computational Linguistics}, pages 311--318.

\bibitem[{Sahoo et~al.(2024{\natexlab{a}})Sahoo, Meharia, Ghosh, Saha, Jain, and Chadha}]{sahoo2024unveiling}
Pranab Sahoo, Prabhash Meharia, Akash Ghosh, Sriparna Saha, Vinija Jain, and Aman Chadha. 2024{\natexlab{a}}.
\newblock \href {http://arxiv.org/abs/2405.09589} {Unveiling hallucination in text, image, video, and audio foundation models: A comprehensive survey}.

\bibitem[{Sahoo et~al.(2024{\natexlab{b}})Sahoo, Singh, Saha, Jain, Mondal, and Chadha}]{sahoo2024systematic}
Pranab Sahoo, Ayush~Kumar Singh, Sriparna Saha, Vinija Jain, Samrat Mondal, and Aman Chadha. 2024{\natexlab{b}}.
\newblock A systematic survey of prompt engineering in large language models: Techniques and applications.
\newblock \emph{arXiv preprint arXiv:2402.07927}.

\bibitem[{Sarker and Gonzalez(2015)}]{sarker2015portable}
Abeed Sarker and Graciela Gonzalez. 2015.
\newblock Portable automatic text classification for adverse drug reaction detection via multi-corpus training.
\newblock \emph{Journal of biomedical informatics}, 53:196--207.

\bibitem[{Sarker et~al.(2016)Sarker, Nikfarjam, and Gonzalez}]{sarker2016social}
Abeed Sarker, Azadeh Nikfarjam, and Graciela Gonzalez. 2016.
\newblock Social media mining shared task workshop.
\newblock In \emph{Biocomputing 2016: Proceedings of the Pacific Symposium}, pages 581--592. World Scientific.

\bibitem[{Sato et~al.(2022)Sato, Yamada, and Kashima}]{sato2022re}
Ryoma Sato, Makoto Yamada, and Hisashi Kashima. 2022.
\newblock Re-evaluating word mover’s distance.
\newblock In \emph{International Conference on Machine Learning}, pages 19231--19249. PMLR.

\bibitem[{Simonyan and Zisserman(2014)}]{simonyan2014very}
Karen Simonyan and Andrew Zisserman. 2014.
\newblock Very deep convolutional networks for large-scale image recognition.
\newblock \emph{arXiv preprint arXiv:1409.1556}.

\bibitem[{Sultana et~al.(2013)Sultana, Cutroneo, and Trifir{\`o}}]{sultana2013clinical}
Janet Sultana, Paola Cutroneo, and Gianluca Trifir{\`o}. 2013.
\newblock Clinical and economic burden of adverse drug reactions.
\newblock \emph{Journal of Pharmacology and Pharmacotherapeutics}, 4(1\_suppl):S73--S77.

\bibitem[{Thawkar et~al.(2023)Thawkar, Shaker, Mullappilly, Cholakkal, Anwer, Khan, Laaksonen, and Khan}]{thawkar2023xraygpt}
Omkar Thawkar, Abdelrahman Shaker, Sahal~Shaji Mullappilly, Hisham Cholakkal, Rao~Muhammad Anwer, Salman Khan, Jorma Laaksonen, and Fahad~Shahbaz Khan. 2023.
\newblock Xraygpt: Chest radiographs summarization using medical vision-language models.
\newblock \emph{arXiv preprint arXiv:2306.07971}.

\bibitem[{Tutubalina et~al.(2017)Tutubalina, Nikolenko et~al.}]{tutubalina2017combination}
Elena Tutubalina, Sergey Nikolenko, et~al. 2017.
\newblock Combination of deep recurrent neural networks and conditional random fields for extracting adverse drug reactions from user reviews.
\newblock \emph{Journal of healthcare engineering}, 2017.

\bibitem[{Viera et~al.(2005)Viera, Garrett et~al.}]{viera2005understanding}
Anthony~J Viera, Joanne~M Garrett, et~al. 2005.
\newblock Understanding interobserver agreement: the kappa statistic.
\newblock \emph{Fam med}, 37(5):360--363.

\bibitem[{Wang et~al.(2022)Wang, Yang, Hu, Li, Lin, Gan, Liu, Liu, and Wang}]{wang2022git}
Jianfeng Wang, Zhengyuan Yang, Xiaowei Hu, Linjie Li, Kevin Lin, Zhe Gan, Zicheng Liu, Ce~Liu, and Lijuan Wang. 2022.
\newblock Git: A generative image-to-text transformer for vision and language.
\newblock \emph{arXiv preprint arXiv:2205.14100}.

\bibitem[{Wang et~al.(2009)Wang, Hripcsak, Markatou, and Friedman}]{wang2009active}
Xiaoyan Wang, George Hripcsak, Marianthi Markatou, and Carol Friedman. 2009.
\newblock Active computerized pharmacovigilance using natural language processing, statistics, and electronic health records: a feasibility study.
\newblock \emph{Journal of the American Medical Informatics Association}, 16(3):328--337.

\bibitem[{Yadav et~al.(2018{\natexlab{a}})Yadav, Ekbal, and Saha}]{yadav2018feature}
Shweta Yadav, Asif Ekbal, and Sriparna Saha. 2018{\natexlab{a}}.
\newblock Feature selection for entity extraction from multiple biomedical corpora: A pso-based approach.
\newblock \emph{Soft Computing}, 22(20):6881--6904.

\bibitem[{Yadav et~al.(2018{\natexlab{b}})Yadav, Ekbal, Saha, Bhattacharyya, and Sheth}]{yadav2018multi}
Shweta Yadav, Asif Ekbal, Sriparna Saha, Pushpak Bhattacharyya, and Amit Sheth. 2018{\natexlab{b}}.
\newblock Multi-task learning framework for mining crowd intelligence towards clinical treatment.

\bibitem[{Yadav et~al.(2020)Yadav, Ramesh, Saha, and Ekbal}]{yadav2020relation}
Shweta Yadav, Srivatsa Ramesh, Sriparna Saha, and Asif Ekbal. 2020.
\newblock Relation extraction from biomedical and clinical text: Unified multitask learning framework.
\newblock \emph{IEEE/ACM transactions on computational biology and bioinformatics}, 19(2):1105--1116.

\bibitem[{Zhang et~al.(2023)Zhang, Huang, Jin, and Lu}]{zhang2023visionlanguage}
Jingyi Zhang, Jiaxing Huang, Sheng Jin, and Shijian Lu. 2023.
\newblock \href {http://arxiv.org/abs/2304.00685} {Vision-language models for vision tasks: A survey}.

\bibitem[{Zhang et~al.(2019)Zhang, Kishore, Wu, Weinberger, and Artzi}]{zhang2019bertscore}
Tianyi Zhang, Varsha Kishore, Felix Wu, Kilian~Q Weinberger, and Yoav Artzi. 2019.
\newblock Bertscore: Evaluating text generation with bert.
\newblock \emph{arXiv preprint arXiv:1904.09675}.

\bibitem[{Zhang et~al.(2016)Zhang, Nie, and Zhang}]{zhang2016ensemble}
Zhifei Zhang, JY~Nie, and Xuyao Zhang. 2016.
\newblock An ensemble method for binary classification of adverse drug reactions from social media.
\newblock In \emph{Proceedings of the Social Media Mining Shared Task Workshop at the Pacific Symposium on Biocomputing}, volume~1.

\bibitem[{Zhao et~al.(2019)Zhao, Peyrard, Liu, Gao, Meyer, and Eger}]{zhao2019moverscore}
Wei Zhao, Maxime Peyrard, Fei Liu, Yang Gao, Christian~M Meyer, and Steffen Eger. 2019.
\newblock Moverscore: Text generation evaluating with contextualized embeddings and earth mover distance.
\newblock \emph{arXiv preprint arXiv:1909.02622}.

\bibitem[{Zhou and Gao(2023)}]{zhou2023skingpt}
Juexiao Zhou and Xin Gao. 2023.
\newblock Skingpt: A dermatology diagnostic system with vision large language model.
\newblock \emph{arXiv preprint arXiv:2304.10691}.

\end{thebibliography}
\bibliographystyle{acl_natbib}

\section*{Appendix}
\label{sec:appendixA}

\section{Experimental Settings}
\label{es}
This section provides the hyperparameters and experimental setups utilized in the study. All experiments were conducted using multiple RTX 2080Ti GPUs. The dataset was partitioned, allocating 80\% for training and 20\% for testing. For InstructBLIP, the learning rate was set to 1e-5, executed for 50 epochs, with a batch size of 2. Similarly, for the BLIP model, a learning rate of 0.0001 was utilized and executed for 50 epochs, with a batch size of 2. For GIT fine-tuning, a learning rate of 5e-3 was applied for 50 epochs, with a batch size of 2. All models were implemented using Scikit-Learn~\footnote{\url{https://scikit-learn.org/stable/}} and PyTorch as the backend framework~\footnote{\url{https://pytorch.org/}}. 

\section{Fine-tuning}
\label{FT}
We have followed several steps to fine-tune BLIP on our multimodal dataset.
First, we have prepared the dataset in JSON format, which is compatible with the BLIP framework, and each image-text pair is represented as a dictionary with the following keys: image\_path: The path to the image file, text: The caption or other text description of the image. We have fine-tuned BLIP with \texttt{learning rate = 0.001}, \texttt{number of epochs = 50}, and \texttt{batch size = 16}. The training process takes 6 hours for the fine-tuning, and we evaluate the performance on a held-out test set.

To fine-tune the GIT model on the proposed MMADE dataset, we utilized the Hugging Face Transformers library and followed a systematic process. First, we prepared the data using a PyTorch dataset, converting it into the required format using the \texttt{GitProcessor} class. Finally, we have utilized the GIT-base model, which is pre-trained on a substantial dataset of image-text pairs. We have utilized Parameters such as a \texttt{learning rate = 5e-3, epochs = 30,} and \texttt{batch size = 2}. The optimization was performed using the Adam optimizer with the cross-entropy loss function, and the fine-tuning process took 3 hours.

To fine-tune the CNN-LSTM model, we leverage the CNN models (VGG16 and ResNet50) to extract image features, followed by an LSTM for sequence generation. Model compilation involves the use of categorical cross-entropy loss and the Adam optimizer. Training proceeds for 40 epochs, utilizing a batch size of 32.



\section{Evaluation Metrics}
\label{em}
We have utilized BLEU score~\cite{papineni2002bleu}, ROUGE score~\cite{lin2004rouge}, BERTScore~\cite{zhang2019bertscore} and MoverScore~\cite{zhao2019moverscore}. BLEU score is utilized to assess the quality of machine-generated text by comparing it to human-generated reference text. It measures the similarity in word sequences between the machine-generated and human reference texts using n-grams, penalizing shorter machine-generated texts to provide a quantitative measure of translation accuracy. The ROUGE score is also used to evaluate the quality of machine-generated summaries compared to human-written summaries. It works by calculating the overlap of n-grams between the machine-generated summary and the reference summaries. ROUGE and BLEU metrics evaluate text quality based on syntactic overlap, considering unigrams and bigrams, lacking the ability to decode semantic meaning effectively. However, BERTScore focuses on understanding the semantic meaning of generated text compared to the intended text, enabling a more nuanced and accurate comparison. MoverScore is a way to measure how similar a machine-generated text is to a human-written text. It does this by using BERT embeddings to understand the meaning of the words and sentences in both texts and then using Word Mover's Distance~\cite{sato2022re} to measure how similar the two texts are.

\section{Statistical Analysis}
\label{SA}

In this study, we employed the paired t-test to assess the statistical significance of differences between the outcomes of unimodal and multimodal models. The null hypothesis (H0) assumes no significant disparity in scores, while the alternative hypothesis (H1) proposes the opposite. We structured our analysis around the assumption of paired data, where each BLEU score corresponds to the same model in different settings (Multimodal Vs. Unimodal). By applying the paired t-test, we aimed to rigorously evaluate performance disparities and offer insights into the models' relative effectiveness. The obtained p-values for ROUGE, BLEU and BERTScore scores were 0.008, 0.0097 and 0.0316, respectively, both below the 0.05 threshold, leading to the rejection of the null hypothesis.

\begin{figure}
  \centering
  \includegraphics[width=\linewidth]{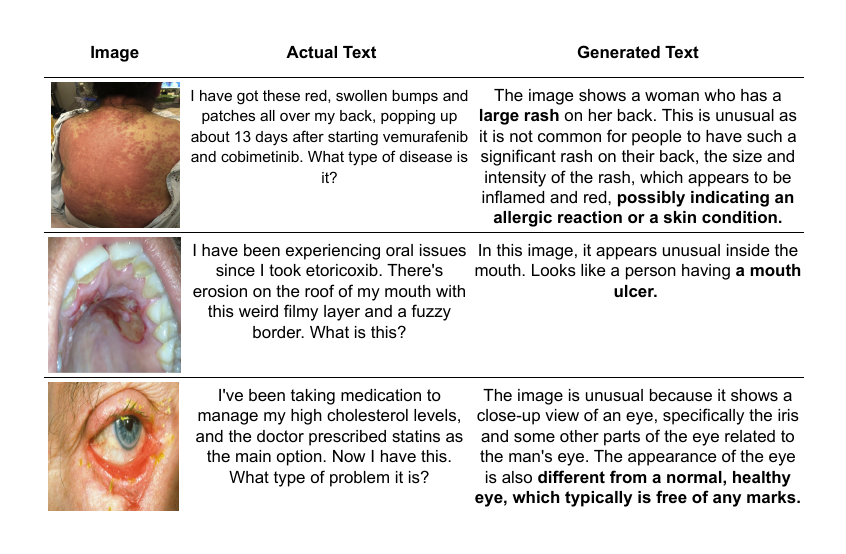}
  \caption{Illustration of a case study showcasing the performance of the InstructBLIP model from the dataset}
  \label{casestudy}
\end{figure}

\begin{figure}[h]
  \centering
  \includegraphics[width=0.8\linewidth]{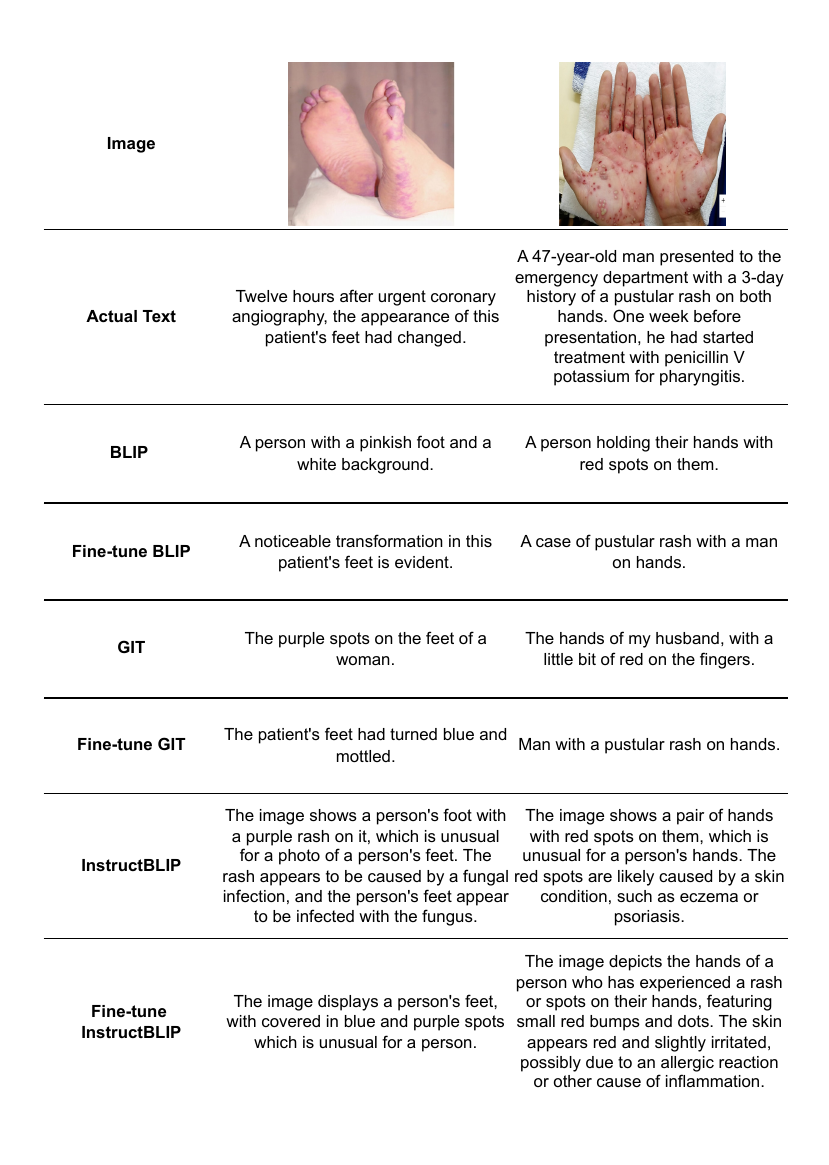}
  \caption{Example of the sample image and corresponding text from the dataset along with generated text from the Proposed model with two other models.}
  \label{outcome}
\end{figure}

\section{Case study}
\label{cs}
The case study depicted in Fig.~\ref{casestudy} reveals intriguing insights across various body parts. In the first row, depicting rashes, fine-tuned InstructBLIP accurately identifies distinct body parts and key findings, potentially attributed to the diverse visual characteristics of different rash types prevalent in the dataset with the largest portion of 81.06\%. However, performance in identifying mouth-related ADEs is less satisfactory despite comprising 5.93\% of the dataset, likely due to potential confusion with other oral features like the tongue, teeth, or lips. Conversely, despite eye-related problems representing only 1.8\% of the dataset, the model performs comparatively better in this category, possibly because it focuses specifically on infected regions, enhancing its ability to identify relevant features accurately.

\section{Comparative output}
\label{co}
We have added two more examples in Fig.~\ref{outcome} showing the ADR instances, corresponding ground truth, and the model-generated outputs.

\end{document}